\pdfoutput=1
\documentclass{article} 
\usepackage[margin=1in]{geometry}
\usepackage{times}
\usepackage[round]{natbib}
\let\cite\citep

\usepackage{hyperref}
\usepackage{graphicx}
\usepackage{url}
\usepackage{amsmath,amssymb}

\title{More Experts, Worse Dynamics: \\ Inverse Scaling and Spectral Bias in Mixture-of-Experts State-Space Models}

\author{Chandresh Pandey \\
Computer Science and Engineering, Delhi Technological University, India \\
\texttt{chandreshpandey.work@gmail.com}
}
\date{}

\begin{document}

\maketitle

\begin{abstract}
Mixture-of-Experts (MoE) architectures are commonly motivated as a way to increase expressivity by decomposing complex systems into simpler local dynamics. This intuition has recently been extended to spectral state-space models, where mixing stable operators is assumed to enable adaptation to heterogeneous or regime-switching time series. We critically evaluate this assumption in a controlled synthetic setting designed to isolate dynamical rather than representational challenges. We study a next-step prediction task on sequences composed of three regimes: chaotic dynamics generated by the Mackey--Glass system, a stable oscillatory regime, and a noise-dominated autoregressive regime. Across extensive ablations including capacity scaling, oracle routing, frozen-expert variants, and comparisons to output-level MoE baselines, operator-level mixture models consistently fail to outperform a single-expert baseline. Increasing the number of experts leads to inverse scaling, routing collapses or fails to induce meaningful specialization, and even perfect regime supervision does not prevent degradation in global performance. Furthermore, we show that apparent improvements in mean squared error on chaotic trajectories can be misleading. Phase-space analysis reveals that lower error often arises from temporal smoothing that destroys the geometry of the underlying attractor rather than from faithful modeling of the dynamics. These results identify a likely limitation of operator interpolation \emph{under the studied parameterization and training protocol}, and underscore the need for geometry-aware evaluation when assessing regime-switching dynamical systems.
\end{abstract}

\section{Setting and Hypothesis}

Many real-world time series exhibit heterogeneous dynamical regimes, where the underlying system alternates between qualitatively different behaviors such as chaos, periodic oscillation, and stochastic noise. Examples arise in domains including physiological monitoring, climate dynamics, and control systems. In such settings, predictive accuracy alone is insufficient: in regimes with sensitive dependence on initial conditions, preserving the qualitative structure of the dynamics is equally important.

A common modeling assumption is that these systems can be decomposed into a small number of simpler local dynamics, each governing a subset of timesteps. Mixture-of-Experts (MoE) architectures are often proposed \cite{shazeer2017, Eigen2013LearningFR} as a natural consequence of this assumption, with a learned router selecting among specialized experts to adapt to regime changes. Recently, this idea has been extended to spectral state-space models \cite{pióro2024moemamba, anthony2024blackmamba, zhan2025routing, gu2022efficiently}, where mixing multiple stable operators is assumed to increase expressivity while retaining stability and long-range modeling capabilities.

In this work, we test this assumption in a controlled synthetic setting designed to isolate \emph{dynamical} rather than representational challenges. We study a next-step prediction task on sequences composed of three sequential regimes: a chaotic regime generated by the Mackey--Glass system with delay $\tau = 17$, a stable oscillatory regime, and a noise-dominated autoregressive regime. Performance is evaluated using normalized mean squared error (NMSE), both globally and per regime, and supplemented with delay-embedded phase-space analysis to assess dynamical fidelity.

Under the MoE hypothesis, increasing the number of experts should reduce NMSE, induce regime-specific specialization, and preserve the geometry of the chaotic attractor. We treat the approach as unsuccessful if increased capacity fails to improve performance, routing collapses or fails to separate regimes, or apparent gains arise from smoothing that distorts the underlying dynamics.

\section{Proposed Solution}

We evaluate a mixture-based extension of linear state-space models intended to model regime-switching dynamics through expert specialization. The model, which we refer to as \emph{Gated Multi-Stability} (GMS), combines multiple stable state-space operators using a learned routing mechanism.

Each expert in GMS is a linear state-space model parameterized in the spectral domain, with its state transition operator constrained to have eigenvalues strictly inside the unit circle \cite{gu2024mamba}. This constraint enforces stability and prevents unbounded growth, a common design choice in recent spectral state-space models. At each timestep, a router network produces soft weights over experts based on the current input, and the effective transition operator is formed as a convex combination of expert operators.

This formulation assumes that heterogeneous dynamics can be represented as interpolations between a small set of stable local operators, and that soft routing is sufficient to induce specialization across regimes. Under this assumption, increasing the number of experts should improve expressivity while preserving stability, relative to a single-expert baseline. We evaluate this approach under identical training conditions against a single-expert spectral state-space model, and further analyze its behavior through oracle routing, frozen-expert ablations, and comparisons to output-level mixture-of-experts models.

\section{Observed Outcome}
\label{sec:outcome}

We now report the empirical behavior of operator-level mixture models under controlled regime-switching dynamics, focusing on performance scaling, routing behavior, and dynamical fidelity.

\subsection{Global Performance and Capacity Scaling}

\begin{figure}[h]
\begin{center}
\includegraphics[width=0.9\linewidth]{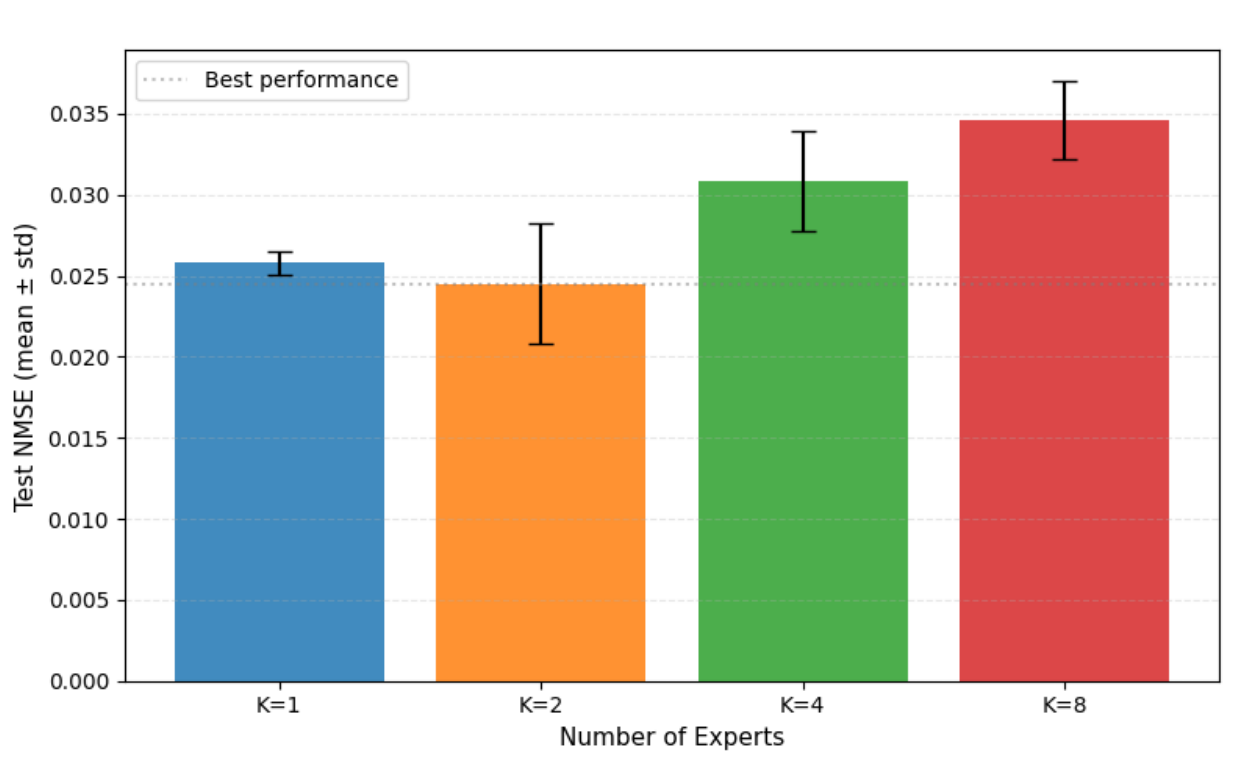}
\end{center}
\caption{Test NMSE (mean $\pm$ standard deviation over five seeds) as a function of the number of experts $K$. A small improvement is occasionally observed for $K=2$, but performance degrades and variance increases as capacity grows beyond a single expert.}
\label{fig:capacity_ablation}
\end{figure}

We examine whether increasing the number of experts improves predictive performance. Figure~\ref{fig:capacity_ablation} shows test NMSE (averaged over five seeds) as a function of the number of experts $K$.

Increasing model capacity does not lead to consistent improvements over the single-expert baseline.
While a modest gain is sometimes observed for $K=2$, models with higher capacity consistently underperform.
In particular, both $K=4$ and $K=8$ exhibit higher mean NMSE and substantially increased variance across seeds.

This behavior is stable across random initializations.
The single-expert model achieves reliable performance with low variance, whereas multi-expert models are more sensitive to initialization and optimization noise.
Notably, the $K=4$ configuration, which serves as the primary setting in subsequent experiments, performs worse on average than the baseline despite having greater representational capacity.

Overall, these results indicate that operator-level mixture models do not benefit from increased capacity in this setting.
Rather than enabling effective specialization, additional experts introduce instability and degrade generalization.
This points to a limitation of operator mixing itself, rather than a failure of training or optimization.

\subsection{Routing Behavior and Regime Specialization}

To understand why increased capacity fails to improve performance, we analyze the routing behavior and its ability to induce expert specialization across regimes. Despite the availability of multiple experts, the router consistently assigns the majority of timesteps to a single expert.

Under standard training with a supervised warmup, the $K=4$ GMS model effectively uses only one expert across all regimes. Chaotic, oscillatory, and noise-dominated segments are routed to the same expert, resulting in only $1/3$ experts being active. This behavior persists after the supervision signal is annealed to zero and training continues under the reconstruction objective alone.

The lack of specialization is reflected in the per-regime error profile. While the GMS model reduces error on the chaotic regime relative to the baseline, it performs worse on the oscillatory regime and shows no improvement on the noise regime. These regressions offset the localized gains, leading to a slight degradation in global NMSE.

\begin{figure}[h]
\centering
\includegraphics[
  width=\linewidth,
  trim=0 0 0 282,
  clip
]{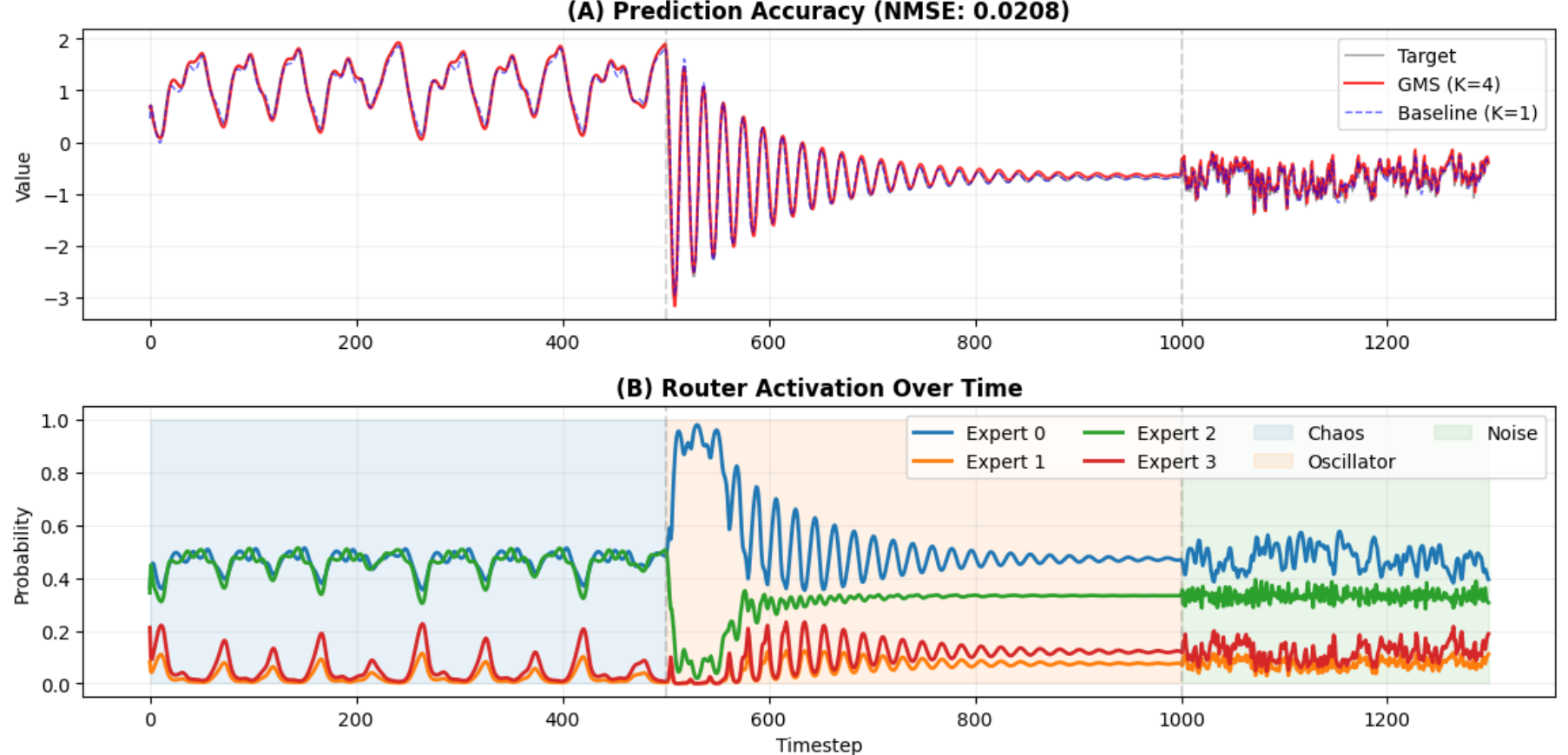}
\caption{Prediction and routing behavior across regimes.
\textbf{Top:} One-step prediction for the single-expert baseline ($K=1$) and the GMS model ($K=4$) across chaotic, oscillatory, and noise-dominated segments.
\textbf{Bottom:} Router assignment probabilities over time for the $K=4$ GMS model. Despite explicit regime boundaries, routing collapses to a dominant expert \cite{chi2022on,dai-etal-2022-stablemoe} for most timesteps, with only transient deviations during regime transitions.}
\label{fig:routing_dynamics}
\end{figure}

Figure~\ref{fig:routing_dynamics} illustrates this collapse directly. Even with clear regime boundaries, the router overwhelmingly favors a single expert throughout training, preventing meaningful expert specialization. This suggests that soft routing alone is insufficient to induce regime separation when experts are mixed at the operator level.

\subsection{Phase-Space Analysis: Geometry versus MSE}

Normalized mean squared error (NMSE) is a standard metric for time-series prediction, but it is often inadequate for evaluating chaotic systems. In such regimes, small phase errors accumulate rapidly, leading to large pointwise deviations even when the qualitative dynamics are preserved. Conversely, models that suppress variability can achieve deceptively low MSE while failing to capture the underlying structure of the system.

To assess whether improvements in NMSE reflect genuine dynamical modeling, we analyze delay-coordinate phase-space reconstructions of the chaotic regime \cite{10.1007/BFb0091924}. Specifically, for trajectories generated by the Mackey--Glass system with delay $\tau = 17$, we reconstruct the embedding $(x(t), x(t-\tau))$ and compare ground truth dynamics to model predictions \cite{doi:10.1126/science.267326}.

\begin{figure}[h]
  \centering
  \includegraphics[width=\linewidth]{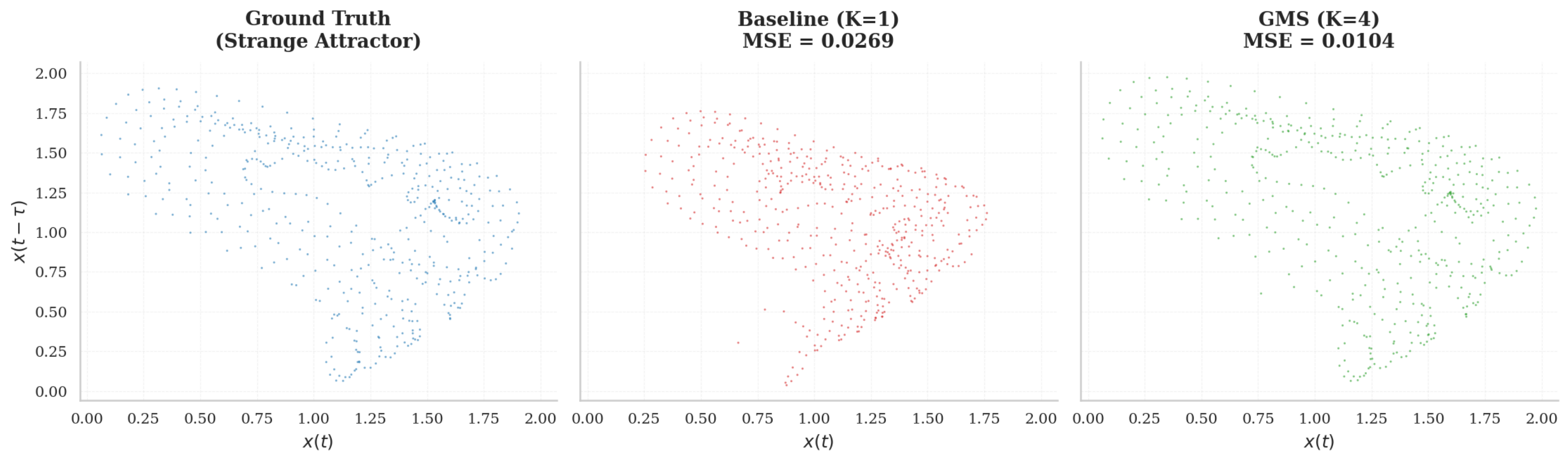}
  \caption{Phase-space reconstruction of the chaotic regime using delay-coordinate embedding $(x(t), x(t-\tau))$ with $\tau = 17$.
  \textbf{Left}: ground truth Mackey--Glass attractor.
  \textbf{Middle}: single-expert baseline ($K=1$).
  \textbf{Right}: GMS with operator mixing ($K=4$).
  Despite a $61\%$ reduction in MSE, operator mixing collapses the attractor toward a smooth, low-variance manifold, destroying fine-scale chaotic structure.}
  \label{fig:phase_space}
\end{figure}

Figure~\ref{fig:phase_space} reveals a mismatch between error metrics and dynamical fidelity. While the single-expert baseline partially preserves the folded geometry of the Mackey--Glass attractor, the GMS model produces a strongly contracted trajectory collapsing toward a smooth manifold. Although this yields lower MSE, it eliminates fine-scale chaotic structure. The apparent improvement arises from temporal smoothing and variance suppression rather than faithful reproduction of the dynamics, indicating a central failure mode of operator mixing. Ablations with oracle routing and frozen experts (Appendix~D) confirm this persists even under idealized settings.

\section{Conclusion}

We investigated whether operator-level Mixture-of-Experts can improve modeling of regime-switching dynamical systems when combined with spectral state-space models. Despite strong intuitive motivation and extensive ablations, the proposed operator-mixing formulation consistently failed to outperform a single-expert baseline. Increasing capacity led to inverse scaling \cite{mckenzie2023inverse}, routing collapsed or failed to induce useful specialization, and even oracle supervision did not resolve the degradation.

Crucially, apparent improvements in standard error metrics can be misleading. Phase-space analysis shows that lower MSE in chaotic regimes often arises from temporal smoothing that destroys the geometry of the underlying attractor, indicating that optimization favors stability and contraction over faithful reproduction of the dynamics \cite{churchill2022deeplearningchaoticsystems}. Overall, our results suggest that, \emph{for the convex spectral-operator parameterization studied}, interpolation of stable operators biases learned dynamics toward averaged, contractive behavior. More broadly, naive expert interpolation can suppress regime-specific structure rather than enhance it, highlighting the need for geometry-aware evaluation in spectral SSMs. \textbf{Remark.}
A formal analysis of this contraction is given in Appendix~\ref{app:theory_scope}.

\bibliography{arxiv2026}

@inproceedings{
shazeer2017,
title={Outrageously Large Neural Networks: The Sparsely-Gated Mixture-of-Experts Layer},
author={Noam Shazeer and Azalia Mirhoseini and Krzysztof Maziarz and Andy Davis and Quoc Le and Geoffrey Hinton and Jeff Dean},
booktitle={International Conference on Learning Representations},
year={2017},
url={https://openreview.net/forum?id=B1ckMDqlg}
}

@article{Eigen2013LearningFR,
  title={Learning Factored Representations in a Deep Mixture of Experts},
  author={David Eigen and Marc'Aurelio Ranzato and Ilya Sutskever},
  journal={CoRR},
  year={2013},
  volume={abs/1312.4314},
  url={https://api.semanticscholar.org/CorpusID:11492613}
}

@inproceedings{
pióro2024moemamba,
title={{MoE-Mamba}: Efficient Selective State Space Models with Mixture of Experts},
author={Maciej Pi{\'o}ro and Kamil Ciebiera and Krystian Kr{\'o}l and Jan Ludziejewski and Micha{\l} Krutul and Jakub Krajewski and Szymon Antoniak and Piotr Mi{\l}o{\'s} and Marek Cygan and Sebastian Jaszczur},
booktitle={ICLR 2024 Workshop on Mathematical and Empirical Understanding of Foundation Models},
year={2024},
url={https://openreview.net/forum?id=LRp8rCaYH7}
}

@inproceedings{
anthony2024blackmamba,
title={{BlackMamba}: Mixture of Experts for State-Space Models},
author={Quentin Gregory Anthony and Yury Tokpanov and Paolo Glorioso and Beren Millidge},
booktitle={ICLR 2024 Workshop on Mathematical and Empirical Understanding of Foundation Models},
year={2024},
url={https://openreview.net/forum?id=10dsmPgq9L}
}

@inproceedings{
zhan2025routing,
title={{Routing Mamba}: Scaling State Space Models with Mixture-of-Experts Projection},
author={Zheng Zhan and Liliang Ren and Shuohang Wang and Liyuan Liu and Yang Liu and Yeyun Gong and Yanzhi Wang and Yelong Shen},
booktitle={The Thirty-ninth Annual Conference on Neural Information Processing Systems},
year={2025},
url={https://openreview.net/forum?id=lqywifxoo1}
}

@inproceedings{
gu2022efficiently,
title={Efficiently Modeling Long Sequences with Structured State Spaces},
author={Albert Gu and Karan Goel and Christopher Re},
booktitle={International Conference on Learning Representations},
year={2022},
url={https://openreview.net/forum?id=uYLFoz1vlAC}
}

@misc{
gu2024mamba,
title={Mamba: Linear-Time Sequence Modeling with Selective State Spaces},
author={Albert Gu and Tri Dao},
year={2024},
url={https://openreview.net/forum?id=AL1fq05o7H}
}

@article{
doi:10.1126/science.267326,
author = {Michael C. Mackey  and Leon Glass },
title = {Oscillation and Chaos in Physiological Control Systems},
journal = {Science},
volume = {197},
number = {4300},
pages = {287-289},
year = {1977},
doi = {10.1126/science.267326},
URL = {https://www.science.org/doi/abs/10.1126/science.267326},
eprint = {https://www.science.org/doi/pdf/10.1126/science.267326}}

@InProceedings{10.1007/BFb0091924,
author="Takens, Floris",
editor="Rand, David
and Young, Lai-Sang",
title="Detecting strange attractors in turbulence",
booktitle="Dynamical Systems and Turbulence, Warwick 1980",
year="1981",
publisher="Springer Berlin Heidelberg",
address="Berlin, Heidelberg",
pages="366--381",
isbn="978-3-540-38945-3"
}

@inproceedings{
chi2022on,
title={On the Representation Collapse of Sparse Mixture of Experts},
author={Zewen Chi and Li Dong and Shaohan Huang and Damai Dai and Shuming Ma and Barun Patra and Saksham Singhal and Payal Bajaj and Xia Song and Xian-Ling Mao and Heyan Huang and Furu Wei},
booktitle={Advances in Neural Information Processing Systems},
editor={Alice H. Oh and Alekh Agarwal and Danielle Belgrave and Kyunghyun Cho},
year={2022},
url={https://openreview.net/forum?id=mWaYC6CZf5}
}

@inproceedings{dai-etal-2022-stablemoe,
    title = "{S}table{M}o{E}: Stable Routing Strategy for Mixture of Experts",
    author = "Dai, Damai  and
      Dong, Li  and
      Ma, Shuming  and
      Zheng, Bo  and
      Sui, Zhifang  and
      Chang, Baobao  and
      Wei, Furu",
    editor = "Muresan, Smaranda  and
      Nakov, Preslav  and
      Villavicencio, Aline",
    booktitle = "Proceedings of the 60th Annual Meeting of the Association for Computational Linguistics (Volume 1: Long Papers)",
    month = may,
    year = "2022",
    address = "Dublin, Ireland",
    publisher = "Association for Computational Linguistics",
    url = "https://aclanthology.org/2022.acl-long.489/",
    doi = "10.18653/v1/2022.acl-long.489",
    pages = "7085--7095",
}

@article{
mckenzie2023inverse,
title={Inverse Scaling: When Bigger Isn't Better},
author={Ian R. McKenzie and Alexander Lyzhov and Michael Martin Pieler and Alicia Parrish and Aaron Mueller and Ameya Prabhu and Euan McLean and Xudong Shen and Joe Cavanagh and Andrew George Gritsevskiy and Derik Kauffman and Aaron T. Kirtland and Zhengping Zhou and Yuhui Zhang and Sicong Huang and Daniel Wurgaft and Max Weiss and Alexis Ross and Gabriel Recchia and Alisa Liu and Jiacheng Liu and Tom Tseng and Tomasz Korbak and Najoung Kim and Samuel R. Bowman and Ethan Perez},
journal={Transactions on Machine Learning Research},
issn={2835-8856},
year={2023},
url={https://openreview.net/forum?id=DwgRm72GQF},
note={Featured Certification}
}

@misc{churchill2022deeplearningchaoticsystems,
      title={Deep Learning of Chaotic Systems from Partially-Observed Data},
      author={Victor Churchill and Dongbin Xiu},
      year={2022},
      eprint={2205.08384},
      archivePrefix={arXiv},
      primaryClass={cs.LG},
      url={https://arxiv.org/abs/2205.08384},
}

\appendix

\section{Training Details and Optimization}
\label{app:training}

This section provides full implementation details for all experiments, including model architecture, optimization settings, training procedures, and supervision schedules. All models were implemented in PyTorch and trained on a single NVIDIA GPU.

\paragraph{Model architecture.}
All models are based on linear spectral state-space operators with complex-valued dynamics. Each expert is parameterized by a diagonal state transition operator in the spectral domain, with eigenvalues constrained to lie strictly inside the unit circle to enforce stability. Unless otherwise stated, all experiments use the same base architecture summarized in Table~\ref{tab:architecture}. Importantly, the state dimension $d_{state}=64$ is defined \emph{per expert}. Thus, for mixture models the total latent dimensionality scales linearly with the number of experts $K$ (e.g., $K=4$ corresponds to 256 latent dimensions), and multi-expert models have strictly larger parameter budgets than the single-expert baseline.

\begin{table}[h]
\centering
\begin{tabular}{l c}
\hline
\textbf{Component} & \textbf{Value} \\
\hline
Input dimension & $d_{in} = 1$ \\
State dimension & $d_{state} = 64$ \\
Output dimension & $d_{out} = 1$ \\
Number of experts & $K \in \{1,2,4,8\}$ \\
State transition & Diagonal spectral operator \\
Eigenvalue type & Complex-valued \\
Stability constraint & $|\lambda_i| < 1$ (via sigmoid) \\
Observation model & Linear readout \\
Nonlinearity & None (fully linear dynamics) \\
\hline
\end{tabular}
\caption{Architectural hyperparameters shared across all models.}
\label{tab:architecture}
\end{table}

\begin{table}[h]
\centering
\begin{tabular}{l c c c}
\hline
\textbf{Experts $K$} & \textbf{$d_{state}$ per expert} & \textbf{Total state dim} & \textbf{Relative capacity} \\
\hline
1 & 64 & 64 & 1.0× \\
2 & 64 & 128 & 2.0× \\
4 & 64 & 256 & 4.0× \\
8 & 64 & 512 & 8.0× \\
\hline
\end{tabular}
\caption{Effective latent dimensionality and relative capacity as a function of number of experts.}
\label{tab:param_budget}
\end{table}

Each expert is parameterized by complex eigenvalues
\[
\lambda_k = r_k \exp(i \theta_k),
\]
where magnitudes $r_k$ are obtained via
\[
r_k = \sigma(\text{logit}_k) \cdot (1 - \epsilon),
\]
with $\epsilon = 0.01$ enforcing strict stability.

Experts are initialized with heterogeneous spectral profiles to encourage diversity. For $K=4$, initial mean magnitudes are approximately $\{0.95, 0.92, 0.65, 0.50\}$, corresponding to high-memory, oscillatory, moderate-decay, and low-memory dynamics respectively. Phases are initialized either linearly or uniformly at random depending on expert index.

\paragraph{Router.}
For mixture models ($K>1$), routing is performed by a feedforward network that maps the current input and the magnitude of the latent state to a distribution over experts. The router observes only $x(t)$ and $|h(t)|$. Its architecture is summarized in Table~\ref{tab:router}.

\begin{table}[h]
\centering
\begin{tabular}{l c}
\hline
\textbf{Router Component} & \textbf{Value} \\
\hline
Expert input projections & $K$ linear layers: $1 \rightarrow 16$ \\
Router input dimension & $16K + 64$ \\
Hidden layer & Linear($16K+64 \rightarrow 4K$) \\
Activation & ReLU \\
Output layer & Linear($4K \rightarrow K$) \\
Normalization & Softmax \\
Temperature & $T = 0.3$ \\
\hline
\end{tabular}
\caption{Router architecture and hyperparameters.}
\label{tab:router}
\end{table}

\paragraph{Dataset.}
All experiments use a synthetic dataset composed of three sequential regimes: (i) a chaotic regime generated by the Mackey--Glass system with delay $\tau=17$, (ii) a stable oscillatory regime generated by a second-order linear oscillator, and (iii) a noise-dominated regime generated by an AR(1) process. Each sequence consists of 500 chaotic steps, followed by 500 oscillatory steps, and 300 noise steps. Signals are normalized to zero mean and unit variance.

\paragraph{Optimization and training.}
All models are trained using identical optimization settings summarized in Table~\ref{tab:optimization}. Training is performed using full-batch gradient descent with AdamW and gradient clipping.

\begin{table}[h]
\centering
\begin{tabular}{l c}
\hline
\textbf{Hyperparameter} & \textbf{Value} \\
\hline
Optimizer & AdamW \\
Learning rate & $1 \times 10^{-3}$ \\
Weight decay & $1 \times 10^{-4}$ \\
Loss function & Mean squared error \\
Gradient clipping & $\ell_2$ norm clipped to $1.0$ \\
Batch size & Full batch \\
Train/test split & 80\% / 20\% \\
Number of epochs & 150 (main), 80 (ablations) \\
\hline
\end{tabular}
\caption{Optimization hyperparameters.}
\label{tab:optimization}
\end{table}

For GMS models, training begins with a supervised routing warmup phase in which the router is trained to predict ground-truth regime labels. The supervision coefficient is annealed linearly over the first 15 epochs:
\[
\alpha(t) = \max\left(0, 1 - \frac{t}{15}\right).
\]
After epoch 15, the router is trained solely via the reconstruction objective. In oracle routing experiments, $\alpha(t)=1$ for all epochs.

The full training objective for GMS models is:
\[
\mathcal{L}
= \mathcal{L}_{\text{MSE}}
+ \alpha \mathcal{L}_{\text{router}}
+ 0.01 \cdot \mathcal{L}_{\text{entropy}}
+ 0.05 \cdot \mathcal{L}_{\text{load}}
+ 0.02 \cdot \mathcal{L}_{\text{diversity}},
\]
where entropy and load regularization encourage balanced routing, and the diversity term penalizes cosine similarity between expert eigenvalue magnitudes.

\paragraph{Reproducibility.}
All experiments are repeated with five random seeds $\{42,100,1337,2024,999\}$. Randomness is controlled via deterministic PyTorch and CUDA settings.

\paragraph{Code and data release.}
We will release a runnable repository with code to reproduce all experiments, the synthetic data generator, and exact random seeds. The repository will include scripts to reproduce Figures~1--4 and the per-seed results reported in Appendix~\ref{app:seeds}.

\section{Theoretical Analysis: Spectral Contraction under Operator Mixing}
\label{app:theory_scope}

We now show that convex combinations of distinct stable spectral operators are inherently contractive, which explains the temporal smoothing and attractor collapse observed empirically.

Consider two diagonal spectral operators
\[
A_1 = \mathrm{diag}(r e^{i\theta_1}), \quad
A_2 = \mathrm{diag}(r e^{i\theta_2}),
\]
with $r < 1$ and $\theta_1 \neq \theta_2$.
A convex mixture produces the effective operator
\[
A_{\text{mix}} = \alpha A_1 + (1-\alpha) A_2,
\quad \alpha \in (0,1).
\]
The corresponding eigenvalue is
\[
\lambda_{\text{mix}} = \alpha r e^{i\theta_1} + (1-\alpha) r e^{i\theta_2}.
\]
By the triangle inequality,
\[
|\lambda_{\text{mix}}|
\leq \alpha |\lambda_1| + (1-\alpha) |\lambda_2|
= r,
\]
with equality only if $\theta_1 = \theta_2$.
When the phases differ, the inequality is strict:
\[
|\lambda_{\text{mix}}| < r.
\]

Thus, mixing two oscillatory operators with different phases produces a strictly more contractive operator. Energy is dissipated by interpolation itself, even if both experts are individually near-unitary. Repeated mixing therefore drives the effective system toward a spiral sink, explaining the empirically observed temporal smoothing and attractor collapse in Figure~\ref{fig:phase_space}.

\section{Seed Sensitivity and Variance}
\label{app:seeds}

To assess robustness and sensitivity to random initialization, all key experiments are repeated across five independent random seeds:
\[
\{42, 100, 1337, 2024, 999\}.
\]
For each seed, we evaluate both the single-expert baseline ($K=1$) and the operator-mixing model ($K=4$) under identical training conditions.

\paragraph{Per-seed results.}
Table~\ref{tab:seed_results} reports test NMSE for each seed. While individual runs occasionally yield competitive performance for the GMS model, the majority of seeds favor the single-expert baseline.

\begin{table}[h]
\centering
\begin{tabular}{c c c}
\hline
\textbf{Seed} & \textbf{Baseline ($K=1$)} & \textbf{GMS ($K=4$)} \\
\hline
42   & 0.02479 & 0.03442 \\
100  & 0.02513 & 0.03480 \\
1337 & 0.02617 & 0.03259 \\
2024 & 0.02650 & 0.02091 \\
999  & 0.02630 & 0.01893 \\
\hline
\end{tabular}
\caption{Test NMSE across five random seeds for the baseline and GMS models.}
\label{tab:seed_results}
\end{table}

\paragraph{Aggregated statistics.}
Mean and standard deviation across seeds are reported in Table~\ref{tab:seed_stats}. The operator-mixing model exhibits substantially higher variance and a worse mean NMSE than the baseline.

\begin{table}[h]
\centering
\begin{tabular}{c c c}
\hline
\textbf{Model} & \textbf{Mean NMSE} & \textbf{Std} \\
\hline
Baseline ($K=1$) & 0.02578 & 0.00069 \\
GMS ($K=4$)      & 0.02833 & 0.00694 \\
\hline
\end{tabular}
\caption{Mean and standard deviation of test NMSE across seeds.}
\label{tab:seed_stats}
\end{table}

\paragraph{Discussion.}
The single-expert baseline exhibits remarkably low variance and consistent performance across all initializations. In contrast, the operator-mixing model displays substantial instability, with a standard deviation an order of magnitude larger than the baseline (0.00694 vs. 0.00069). Notably, the GMS model is capable of outperforming the baseline in specific instances (e.g., Seeds 2024 and 999), indicating that the architecture can achieve high performance under favorable conditions.

However, these localized successes are offset by large regressions in other runs, resulting in a worse mean NMSE overall. This hit-or-miss behavior suggests that the gains are not robust and that performance is highly sensitive to initialization. Given the small sample size ($n=5$), statistical tests (paired t-test $p=0.53$) do not indicate a significant difference at $\alpha=0.05$. Nonetheless, the combination of increased variance and upward drift in mean error provides descriptive empirical evidence of inverse scaling: increasing capacity via operator-level mixing leads to less reliable and less predictable performance.

\begin{figure}[h]
  \centering
  \includegraphics[
    width=\linewidth,
    trim=0 0 0 462,
    clip
  ]{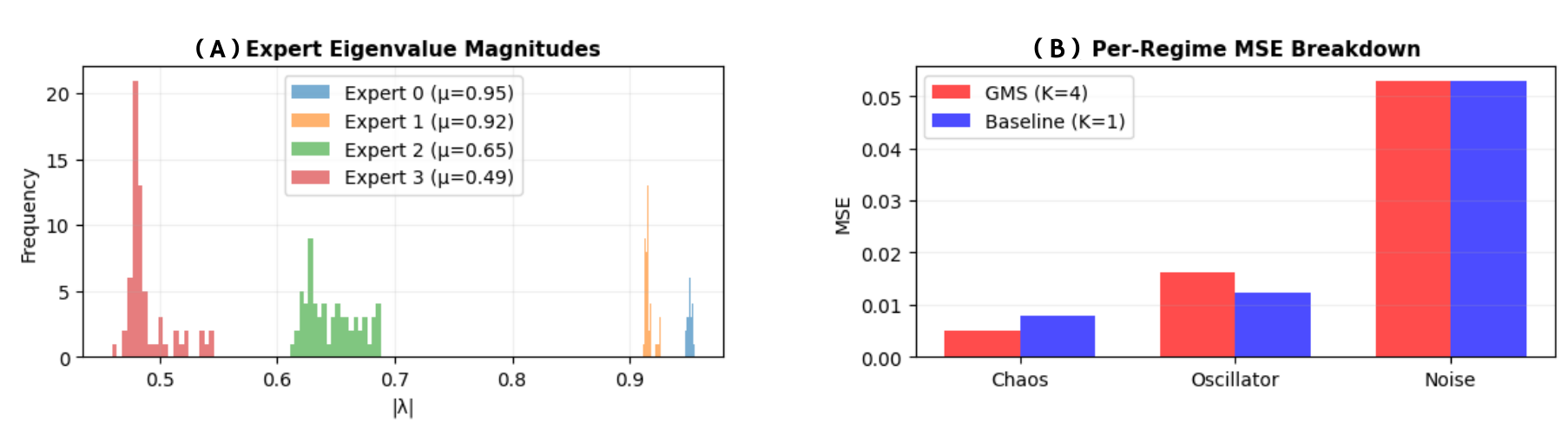}
  \caption{
  (A) Histogram of eigenvalue magnitudes $|\lambda|$ for each expert in the $K=4$ GMS model.
  (B) Per-regime MSE comparison between GMS ($K=4$) and the single-expert baseline ($K=1$).
  Experts differ primarily in degree of contraction, while regime-wise errors reveal a trade-off between improved chaotic prediction and degraded performance on simpler regimes.}
  \label{fig:eigen_hist}
\end{figure}

\paragraph{Spectral structure and regime-wise errors.}
Figure~\ref{fig:eigen_hist} provides additional diagnostics for the $K=4$ GMS model.
Panel (A) shows the distribution of eigenvalue magnitudes for each expert after training. All experts remain strictly contractive, differing mainly in average memory timescales rather than in qualitative spectral structure.
Panel (B) reports per-regime mean squared error, illustrating that operator mixing reduces error in the chaotic regime but degrades performance on the oscillatory and noise regimes, consistent with the trade-offs discussed in the main text.

\section{Oracle Routing and Frozen Expert Ablations}
\label{app:oracle}

A natural hypothesis is that the failure of operator-level mixture models arises from imperfect routing rather than from operator interpolation itself. In this section, we evaluate two ablations designed to isolate the role of routing quality and expert learning.

\paragraph{Oracle routing.}
We first consider an oracle routing variant in which the router is given access to ground-truth regime labels throughout training. The router is trained with full supervision for all epochs and is forced to assign timesteps to the correct expert corresponding to the true regime. This experiment removes routing ambiguity entirely: if operator mixing is fundamentally sound, oracle supervision should enable effective specialization and yield improved global performance.

Table~\ref{tab:oracle_results} reports the resulting test NMSE.

\begin{table}[h]
\centering
\begin{tabular}{l c c}
\hline
\textbf{Model} & \textbf{NMSE} & \textbf{Distinct Experts Used} \\
\hline
Baseline ($K=1$) & 0.02001 & 1/3 \\
GMS Oracle ($K=4$) & 0.02294 & 2/3 \\
\hline
\end{tabular}
\caption{Global NMSE under oracle routing supervision.}
\label{tab:oracle_results}
\end{table}

Despite correct regime assignments and active use of multiple experts, oracle routing leads to \emph{worse} global performance than the single-expert baseline. While error on the chaotic regime is substantially reduced, these gains are offset by regressions on the oscillatory and noise regimes.

\paragraph{Per-regime error.}
To understand this failure mode more precisely, Table~\ref{tab:oracle_regime} reports per-regime MSE under oracle routing.

\begin{table}[h]
\centering
\begin{tabular}{l c c c}
\hline
\textbf{Regime} & \textbf{Baseline ($K=1$)} & \textbf{GMS Oracle ($K=4$)} & $\Delta$ \\
\hline
Chaos & 0.00765 & 0.00265 & +65.3\% \\
Oscillator & 0.01068 & 0.01760 & -64.8\% \\
Noise & 0.05630 & 0.06581 & -16.9\% \\
\hline
\end{tabular}
\caption{Per-regime MSE under oracle routing.}
\label{tab:oracle_regime}
\end{table}

Oracle routing substantially improves performance on the chaotic regime, but simultaneously degrades performance on the easier oscillatory and noise regimes. The net effect is a regression in global NMSE, indicating that operator mixing introduces a structural trade-off between regimes rather than enabling uniform specialization.

\paragraph{Frozen experts.}
We next consider a complementary ablation in which all expert operators are frozen and only the router is trained. In this setting, all spectral parameters $(\lambda_k, B_k, C_k, D_k)$ are fixed at initialization, and the router is trained with full regime supervision to select among these fixed operators.

Table~\ref{tab:frozen_results} reports the resulting performance.

\begin{table}[h]
\centering
\begin{tabular}{l c}
\hline
\textbf{Model} & \textbf{NMSE} \\
\hline
Frozen Experts ($K=4$) & 0.98584 \\
\hline
\end{tabular}
\caption{Global NMSE when expert operators are frozen and only the router is trained.}
\label{tab:frozen_results}
\end{table}

Performance collapses catastrophically, with NMSE increasing by nearly two orders of magnitude. This indicates that routing alone is incapable of compensating for mismatched or suboptimal operators.

\paragraph{Interpretation.}
Taken together, these ablations show that neither improved routing nor decoupling expert learning resolves the failure of operator mixing. Oracle routing demonstrates that even perfect regime supervision does not prevent degradation in global performance, while frozen experts demonstrate that routing alone cannot recover meaningful dynamics.

These results imply that routing collapse observed in standard training is not the root cause of failure, but rather an optimization response to an unfavorable inductive bias. Given the choice between interpolating incompatible operators or collapsing to a single stable one, the optimizer consistently prefers the latter.

\section{Operator Mixing versus Output Mixing}
\label{app:mixing}

The preceding results raise the question of whether the observed degradation is inherent to Mixture-of-Experts architectures in general, or specific to mixing state transition operators. To disentangle these factors, we compare operator-level mixing to a standard output-level MoE under matched conditions.

\paragraph{Experimental setup.}
We evaluate two models with identical expert architectures, routing networks, and optimization settings. In the operator-mixing variant (GMS), the router produces convex weights over expert-specific state transition operators, yielding an effective transition
\[
\Lambda(t) = \sum_{k=1}^K g_k(t)\,\Lambda_k.
\]
In the output-mixing variant (MoE), each expert maintains its own internal dynamics and produces an output prediction, and the router mixes predictions rather than operators:
\[
y(t) = \sum_{k=1}^K g_k\, y_k(t).
\]
In both cases, experts are diagonal spectral state-space models with identical parameterizations, and the same router architecture is used. The only difference is whether interpolation occurs in operator space or output space.

\paragraph{Results.}
Table~\ref{tab:mixing_results} reports test NMSE for both variants.

\begin{table}[h]
\centering
\begin{tabular}{l c}
\hline
\textbf{Model} & \textbf{NMSE} \\
\hline
Operator Mixing (GMS, $K=4$) & 0.08385 \\
Output Mixing (MoE, $K=4$) & 0.07054 \\
\hline
\end{tabular}
\caption{Comparison of operator-level and output-level mixture models.}
\label{tab:mixing_results}
\end{table}

Both mixture models underperform the single-expert baseline, but output mixing is consistently more stable and substantially less degraded than operator mixing. Because the two variants share the same router, training procedure, and expert capacity, this gap isolates the failure to interpolation in operator space rather than to routing capacity or expert expressivity.

\paragraph{Interpretation.}
Operator mixing performs worse because convex combinations of stable state transition operators do not, in general, preserve the qualitative properties of the constituent dynamics. Instead, interpolation produces an averaged system that suppresses regime-specific structure. In contrast, output mixing preserves each expert’s internal dynamics and avoids destructive interference at the level of system evolution.

While output mixing does not outperform the single-expert baseline in this setting, it avoids the catastrophic degradation observed under operator interpolation. These results indicate that the negative outcome is not a rejection of Mixture-of-Experts architectures per se, but a structural limitation of operator-level mixing in spectral state-space models.

\section{LLM Usage Statement}
\label{app:llm}

Large language models were used for language editing and stylistic refinement of the manuscript.
All scientific content, experimental design, implementation, results, and analysis were produced
by the authors.

\end{document}